# Handling Equality Constraints by Adaptive Relaxing Rule for Swarm Algorithms

Xiao-Feng Xie
Institute of Microelectronics,
Tsinghua University,
Beijing 100084, P. R. China
Email:xiexf@ieee.org

Wen-Jun Zhang
Institute of Microelectronics,
Tsinghua University,
Beijing 100084, P. R. China
Email: zwj@tsinghua.edu.cn

De-Chun Bi
Department of Environmental Engineering,
Liaoning University of Petroleum
& Chemical Technology
Fushun, Liaoning, 113008, P. R. China

*Abstract-* The adaptive constraints relaxing rule for swarm algorithms to handle with the problems with eqaulity constraints is presented. The feasible space of such problems may be similar to ridge function class, which is hard for applying swarm algorithms. To enter the solution space more easily, the relaxed quasi feasible space is introduced and shrinked adaptively. The experimental results on benchmark functions are compared with the performance of other algorithms, which show its efficiency.

## I. INTRODUCTION

A problem with equality constraints can be defined as:

$$\begin{cases} \text{Miniminze}: f(\vec{x}) \\ \text{Subject to}: g_j(\vec{x}) \leq 0 \quad (1 \leq j \leq m_0, j \in \mathbb{Z}) \\ \qquad\qquad h_j(\vec{x}) = 0 \quad (m_0+1 \leq j \leq m, j \in \mathbb{Z}) \end{cases} \quad (1)$$

where $\vec{x}=(x_1,...,x_d,...,x_D)$ and the *search space* (*S*) is a *D*-dimensional space bounded by all the parametric constraints $x_d \in [l_d, u_d]$. $f$ is *objective function*, $g_i$ and $h_j$ are inequality and equality *constraints*, respectively. The set of points, which satisfying all the constraints, is denoted as *feasible space* ($S_F$), and then the *infeasible space* is $S_I = \overline{S}_F \cap S$.

Normally, an equality constraint is transformed into an inequality one, i.e. $g_{j+m_0} = |h_j| - \varepsilon_{h,j} \leq 0$, for some small *violation value*s $\varepsilon_h \geq 0$ [4, 6, 7, 13].

Swarm algorithms are behavioral models based on the concepts of social swarm that belong to learning paradigms [5]. Each swarm comprises a society of autonomous agents, which each agent [2] is worked by executing the simple action rules according available information in iterated learning cycles. Existing examples include particle swarm optimization (PSO) [3, 10, 15], differential evolution (DE) [16] and their hybrid [17], etc.

For swarm algorithms, the information is represented by each point $\vec{x} \in S$, which its goodness is evaluated by the *goodness function* $F(\vec{x})$. Suppose for a point $\vec{x}^*$, there exists $F(\vec{x}^*) \leq F(\vec{x})$ for $\forall \vec{x} \in S$, then $\vec{x}^*$ and $F(\vec{x})$ are separately the global optimum point and its value. In general, the goal is to find point(s) that belong to the *solution space* $S_O = \{\vec{x} \in S_F | F_\Delta(\vec{x}) = F(\vec{x}) - F(\vec{x}^*) \leq \varepsilon_O\}$ instead of the $\vec{x}^*$, where $\varepsilon_O$ is a small positive value.

The goodness function $F(\vec{x})$ is encoded by constraint-handling methods that applied on the problem [12]. In order to avoiding the laborious settings for the penalty coefficients [4, 13], and without requiring a starting point in $S_F$ [8], the method that following Deb's criteria [6] is applied on swarm algorithms for the problems with inequality constraints successfully [11, 17]. However, it is still difficulty to deal with equality constraints [8], especially when the *violation value*s are very small [17].

This paper intends to handling with equality constraints for swarm algorithms. In the section 2, the details on swarm algorithms are introduced. Then in the section 3, the ridge function class problem [1, 13] is analysized, which shows that it is hard for swarm algorithms. In the section 4, the basic constraints handling (BCH) rule, which is following Deb's criteria, is introduced. And then the adaptive constraints relaxing (ACR) rule is then studied, since the BCH rule may bring the problem with eqaulity constraints into ridge function class. Then the method is applied to three benchmark functions [12], and the experimental results are compared with those of existing algorithms [7, 13], which illustrate the significant performance improvement.

## II. SWARM ALGORITHMS

In swarm algorithms, each agent is worked in iterated learning cycles. Supposing the number of agents in a swarm is *N*, and the total number of learning cycles is *T*, then at the *t*th ($1 \leq t \leq T, t \in \mathbb{Z}$) learning cycle, each agent is activated in turn, which generates-and-tests a new point based on its own experience and the social sharing information.

For the convenience of discussion, for the *i*th ($1 \leq i \leq N, i \in \mathbb{Z}$) agent, the point with the best goodness value generated in its past learning cycles is defined as $\vec{p}_i^{(t)}$. The point with the best goodness value in the set $\{\vec{p}_i^{(t)} | 1 \leq i \leq N, i \in \mathbb{Z}\}$ is defined as $\vec{g}^{(t)}$, which is often the main source of the social sharing information.

The total evaluation times are $T_E = N \cdot T$.

*A. Particle swarm (PS) agent*

Particle swarm agent, called *particle* [10], generates $\vec{x}^{(t+1)}$ according to the historical experiences of its own and its colleagues, for the *d*th dimension [15]:

$$v_d^{(t+1)} = w \cdot v_d^{(t)} + c_1 \cdot U_\mathbb{R}() \cdot (p_d^{(t)} - x_d^{(t)}) \quad (2)$$
$$+ c_2 \cdot U_\mathbb{R}() \cdot (g_d^{(t)} - x_d^{(t)})$$
$$x_d^{(t+1)} = x_d^{(t)} + v_d^{(t+1)}$$

where $w$ is *inertia weight*, $U_\mathbb{R}()$ is a random real value between 0 and 1. The default parameter values include: $c_1 = c_2 = 2$.

*B. Differential evolution (DE) agent*

Differential evolution agent generates a new point $\vec{k}^{(t+1)}$ for updating its $\vec{p}^{(t+1)}$. At first, it sets $\vec{x}_{DE}^{(t+1)} = \vec{p}^{(t)}$, and $DR = U_\mathbb{Z}(1, D)$ is a random integer value within $[1, D]$. For the $d$th dimension [16, 17], if $U_\mathbb{R}() < CR$ or $d = DR$, then the following rule is applied:

$$x_{DE,d}^{(t+1)} = g_d^{(t)} + SF \cdot \Delta_{N_V,d}^{(t)} \quad (3)$$

where $CR \in [0,1]$ is *crossover factor*, $DR$ ensures the variation at least in one dimension, $SF$ is *scaling factor*. $\vec{\Delta}_{N_V}^{(t)} = \sum_1^{N_V} \vec{\Delta}_1^{(t)}$, and $\vec{\Delta}_1^{(t)} = \vec{p}_{U_\mathbb{Z}(1,N)}^{(t)} - \vec{p}_{U_\mathbb{Z}(1,N)}^{(t)}$ is called *difference vector*. The default parameter values include: $N_V = 2$, $SF = 1/N_V = 0.5$. At last, if $F(\vec{x}_{DE}^{(t+1)}) \le F(\vec{p}^{(t)})$, then it sets $\vec{p}^{(t+1)} := \vec{x}_{DE}^{(t+1)}$.

*C. Combined DEPS agent*

Agent-based modeling provides a natural framework for tuning the rules of each agent [2]. The action rules of DEPS agent [17] is the combination of a DE rule and a PS rule. At the odd $t$, the PS rule is activated for updating its $\vec{v}^{(t+1)}$, $\vec{x}^{(t+1)}$ and $\vec{p}^{(t+1)}$, then at the even $t$, the DE rule is activated for updating its $\vec{p}_i^{(t+1)}$. Notes the $\vec{v}^{(t)}$ and $\vec{x}^{(t)}$ that required by PS rule are not changed by DE rule. However, both rules share with same $\vec{p}^{(t)}$ and $\vec{g}^{(t)}$, which allows them coupling with each other.

## III. RIDGE FUNCTION CLASS

The ridge function class problem may be regarded as extensions of the sphere model breaking its total rotational symmetry in some dimensions of the parameter space [1].

As shown in figure 1, $S_B(\vec{x}_A) = \{F(\vec{x}) \le F(\vec{x}_A)\}$ is an example of ridge function, $S_O$ is the solution space. For a point $\vec{x}_A$, $I_I$ and $I_O$ is the *improvement interval* [13], i.e. the possible *interval* for improving its goodness, and the distance to $S_O$ along with a single dimension, respectively. The significant feature of ridge function is that the ratio $I_I/I_O$ is small due to the large eccentricities.

The action rules of agents in swarm provide a self-adaptive bell-shaped variations [9, 17], which the *gravity center* is between $\vec{p}_i$ and $\vec{g}$, and the *variation strength* is with consensus on the diversity of swarm. As shown in figure 1, $S_V(\vec{x}_A, \vec{x}_B)$ is a typical variation space as $\vec{g} = \vec{x}_A$ and $\vec{p}_i = \vec{x}_B$.

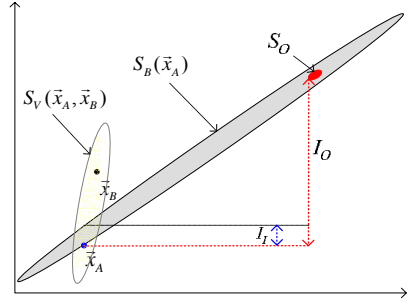

Figure 1. Ridge function class problem.

If an action rule generates new points by modifying only one dimension of current point, then it is obviously that small $I_I/I_O$ results in an increase of the difficulty for convergence, due to the zigzag path in small steps [13].

The common case is that an action rule generates new points by varying several dimensions of current point. It seems that the implicit combination operations by the action rules may accelerate the convergence, as the ($\vec{p} - \vec{g}$) is in suitable direction and enough variation strength. However, as the variation strength is large, it has small possibility for keeping in the direction of ridge axis. Then the ratio $(S_B(\vec{g}) \cap S_V(\vec{p}, \vec{g}))/S_B(\vec{g})$ is very small, which most trails cannot enter $S_B(\vec{g})$. Despite of those invalid trails, the improving $\vec{p}$ often satisfies the short-term goal (increasing goodness by reducing the variation strength) rather than the long-term goal (finding suitable variation strength along the ridge axis for entering the $S_O$). The magnitude of ($\vec{p} - \vec{g}$) is varying to the magnitude of $I_I$ as entering $S_B(\vec{g})$. As $I_I/I_O$ is small, it requires many steps for entering the $S_O$. Moreover, for the bell-shaped variations, the large adjustments are possible, but are much less than the small adjustments. During the many small steps, it is great probability that the agents are clustered near $\vec{g}^{(t)}$ at a certain step and lost the diversity as the large variation are not accepted. Then the variation strength, which is with consensus on the diversity of $\vec{p}$ and $\vec{g}$, becomes small, and then the swarm is easily converged prematurely due to the self-adaptation mechanism.

For problems with equality constraints, if the *violation value*s are very small, then the $S_F$ becomes very narrow. If it is cut into several segments, then some of them have great probability to be in ridge function class landscape. The many directions for the ridge axes of multiple segments increase the difficulty much more.

## IV. BASIC CONSTRAINTS HANDLING (BCH)

The total goodness function $F(\vec{x})$ includes two parts, where $F_{OBJ}(\vec{x}) = f(\vec{x})$ and $F_{CON}(\vec{x}) = \sum_{j=1}^{m} r_j G_j(\vec{x})$ are the goodness functions for objective funtion and constraints, respectively. Here $G_j(\vec{x}) = \max(0, g_j(\vec{x}))$ [4], and $r_j$ are positive weight factors, which default value is equal to 1.

The basic constraints handling (BCH) rule for goodness evaluation is realized by comparing any two points $\vec{x}_A$, $\vec{x}_B$, $F(\vec{x}_A) \leq F(\vec{x}_B)$ when

$$\begin{cases} F_{CON}(\vec{x}_A) < F_{CON}(\vec{x}_B) \text{ OR} \\ F_{OBJ}(\vec{x}_A) \leq F_{OBJ}(\vec{x}_B) \text{ AND } F_{CON}(\vec{x}_A) = F_{CON}(\vec{x}_B) \end{cases} \quad (4)$$

The BCH rule is following Deb's criteria [6]: a) any $\vec{x} \in S_F$ is preferred to any $\vec{x} \notin S_F$; b) among two points in $S_F$, the one having smaller $F_{OBJ}$ is preferred; c) among two points in $S_I$, the one having smaller $F_{CON}$ is preferred.

## V. ADAPTIVE CONSTRAINTS RELAXING (ACR)

For the convenience of discussion, the probability for changing $\vec{x}$ from space $S_X$ to $S_Y$ is defined as $P(S_X \to S_Y)$.

The searching path of the BCH rule is $S_I \to S_F \to S_O$. Then the problems with equality constraints are hard problems for swarm algorithms since the $P(S_F \to S_O)$ can be very small as $S_F$ is in ridge function class.

In this paper, the goodness landscape is transformed by constraints relaxing rule so as to match swarm algorithms. The *quasi feasible space* is defined $S_F' = \{F_{CON}(\vec{x}) \leq \varepsilon_R\}$, where $\varepsilon_R \geq 0$ is the relaxing threshold value, and its corresponding *quasi solution space* is defined as $S_O'$, then an additional rule is applied on equation (4):

$$F_{CON}(\vec{x}) = \max(\varepsilon_R, F_{CON}(\vec{x})) \quad (5)$$

It means that among two points in $S_F'$, the one with smaller $F_{OBJ}$ is preferred. Then the searching path becomes $S_I \to S_F' \to S_O'$. Since $S_F \subseteq S_F'$, it is obviously that $P(S_I \to S_F') \geq P(S_I \to S_F)$. Besides, comparing with $P(S_F \to S_O)$, $P(S_F' \to S_O')$ can increase dramaticly by the enlarged $I_I/I_O$. Then it has $P(S_I \to S_O') \geq P(S_I \to S_O)$.

Of course, $\vec{x} \in S_O'$ does not necessarily mean $\vec{x} \in S_O$. However, the searching path $S_O' \to S_O$ can be built by decreasing $\varepsilon_R$ for increasing $(S_O' \cap S_O)/S_O$. For the extremal case, when $\varepsilon_R = 0$, it has $S_O' = S_O$.

The adjusting of $\varepsilon_R^{(t)}$ is referring to a data repository $\underline{P}$ = $\{\vec{p}_i^{(t)} | 1 \leq i \leq N, i \in \mathbb{Z}\}$, which is updated frequently, normally. In $\underline{P}$, the number of points with $F_{CON}(\vec{p}_i^{(t)}) > \varepsilon_R^{(t)}$ is defined as $N_\varepsilon^{(t)}$.

Initially, the $\varepsilon_R^{(0)}$ is set as maximum $F_{CON}$ value in $K_O$. Then the adaptive relaxing (ACR) rule is employed for ensuring $\varepsilon_R^{(t)} \to 0$, by the following combined rules: a) the basic ratio-keeping sub-rules; b) the forcing sub-rule.

The basic ratio-keeping sub-rules try to keep a balance between the points inside and outside the $S_F'$:

$$\begin{aligned} \text{IF}(r_N^{(t)} \leq r_l) \text{ THEN } \varepsilon_R^{(t+1)} &= \beta_u \cdot \varepsilon_R^{(t)} \\ \text{IF}(r_N^{(t)} \geq r_u) \text{ THEN } \varepsilon_R^{(t+1)} &= \beta_l \cdot \varepsilon_R^{(t)} \end{aligned} \quad (6)$$

where $r_N^{(t)} = N_\varepsilon^{(t)}/N$ is the ratio of the points inside the $S_F'$, $0 \leq r_l < r_u \leq 1$, $0 < \beta_l < 1 < \beta_u$.

The $\varepsilon_R^{(t)}$ is not varied as $r_N^{(t)} \in (r_l, r_u)$. If the $\varepsilon_R^{(t)}$ is large at the end of learning cycles, then $\vec{g}^{(T)}$ may even not enter the $S_O$. Hence it is necessarily to prevent the $\varepsilon_R^{(t)}$ from stagnating for many learning cycles, especially for the cases that the elements in $K_O$ are hardly to be changed.

The forcing sub-rule forces the $\varepsilon_R^{(t)} |_{t \to T} \to 0$:

$$\text{IF}(t \geq t_{th}) \text{ THEN } \varepsilon_R^{(t+1)} = \beta_f \cdot \varepsilon_R^{(t)} \quad (7)$$

where $0 < \beta_f < 1$, and $0 \leq t_{th} \leq T$.

The default parameter values include: $r_l$=0.25, $r_u$=0.75, $\beta_f = \beta_l = 0.618$, $\beta_u = 1.382$, and $t_{th} = 0.5 \cdot T$.

For each learning cycle, the $\vec{g}^{(t)}$ is reevaluated among the $\{\vec{p}_i^{(t)} | 1 \leq i \leq N, i \in \mathbb{Z}\}$, as the $\varepsilon_R^{(t)}$ is updated.

## VI. TEST FUNCTIONS SUITE

Three benchmark functions with equality constraints, called $G_3$, $G_5$, $G_{11}$, $G_{13}$, were described in [12, 13].

### A. $G_3$

Minimum: $f(\vec{x}) = -d^{d/2} \prod_{d=1}^{D} x_d$

subject to: $\sum_{d=1}^{D} x_d^2 - 1 = 0$

where $D$=10 and $0 \leq x_d \leq 1$ $(d = 1, \ldots, D)$.

### B. $G_5$

Minimize:
$f(\vec{x}) = 3\, x_1 + 0.000001\, x_1^3 + 2\, x_2 + 0.000002/3\, x_2^3$
subject to:
$x_4 - x_3 + 0.55 \geq 0$, $\qquad x_3 - x_4 + 0.55 \geq 0$,
$1000\sin(-x_3 - 0.25) + 1000\sin(-x_4 - 0.25) + 984.8 - x_1 = 0$,

$1000\sin(x_3 - 0.25) + 1000\sin(x_3 - x_4 - 0.25) + 984.8 - x_2 = 0$,
$1000\sin(x_4 - 0.25) + 1000\sin(x_4 - x_3 - 0.25) + 1294.8 = 0$.
where $0 \leq x_d \leq 1200$ ($d=1, 2$), $-0.55 \leq x_d \leq 0.55$ ($d=3, 4$).

C. $G_{11}$

Minimize: $f(\vec{x}) = x_1^2 + (x_2 - 1)^2$
subject to: $x_2 - x_1^2 = 0$.
where $-1 \leq x_d \leq 1$ ($d=1, 2$).

D. $G_{13}$

Minimize: $f(\vec{x}) = e^{x_1 x_2 x_3 x_4 x_5}$
subject to: $\sum_{d=1}^{5} x_d^2 - 10 = 0$,
$x_2 x_3 - 5 x_4 x_5 = 0$,
$x_1^3 + x_2^3 + 1 = 0$.
where $-2.3 \leq x_d \leq 2.3$ ($d=1, 2$), $-3.2 \leq x_d \leq 3.2$ ($i=3, 4, 5$).

## VII. RESULTS AND DISCUSSIONS

The violation value determines the difficult of problems. For the larger violation value, which is $\varepsilon_h = 1E-3$, there have already been well solved [11, 17]. Here we only perform experiments on the smaller violation value, which is $\varepsilon_h = 1E-4$, as in the literatures [7, 13].

Table 1 lists the summary of $F^*$ (also is achieved as $\varepsilon_h = 1E-4$) and the mean best fitness value ($F_B$) by two published algorithms: a) (30, 200)-evolution strategy (ES) with stochastic ranking (SR) technique [13], $T=1750$, then $T_E=3.5E5$; and b) genetic algorithm (GA) [7], which $N=70$, $T=2E4$, then $T_E=1.4E6$.

TABLE 1. EXISTING RESULTS FOR TEST PROBLEMS

|       | $G_3$  | $G_5$    | $G_{11}$ | $G_{13}$ |
|-------|--------|----------|----------|----------|
| $F^*$ | 1.0005 | 5126.497 | 0.7499   | 0.053942 |
| ES [13] | 1    | 5128.881 | 0.75     | 0.067543 |
| GA [7]  | 0.9999 | 5432.08 | 0.75     | -        |

Table 2 to 4 summary the mean results by for swarm algorithms in three kinds of constraint-handling methods: a) BCH rule; b) ACR#1, which without the forcing sub-rule; and c) ACR#2, which with the forcing sub-rule, respectively. The values in the parentheses describe the number of runs that are failed in entering $S_F$, and the $F_B$ is calculated by the successful runs only. The boundary constraints are handled by *periodic* mode [17]. The number of agent was set as $N=70$. $CR$ was fixed as 0.9 for the DE rule, and $w$ was fixed as 0.4 for the PS rule to achieve fast convergence. For BCH rule, the learning cycles of all cases were set as $T=5E3$, then $T_E=3.5E5$, which were same as ES in table 1. For ACR rules, for $G_3$, $T=5E3$, then $T_E=3.5E5$; for $G_5$, $G_{11}$ and $G_{13}$, $T=2E3$, then $T_E=1.4E5$. For other algorithm parameters, the default values were used. 100 runs were done for each function.

TABLE 2. RESULTS BY BCH RULE

|      | $G_3$  | $G_5$       | $G_{11}$ | $G_{13}$ |
|------|--------|-------------|----------|----------|
| DE   | 0.3985 | 5133.834    | 0.75056  | 0.52800  |
| PS   | 0.8364 | 5334.97(12) | 0.74992  | 1.10423  |
| DEPS | 0.9849 | 5130.864    | 0.74990  | 0.51065  |

TABLE 3. RESULTS BY ACR#1 RULE
(WITHOUT THE FORCING SUB-RULE)

|      | $G_3$    | $G_5$        | $G_{11}$ | $G_{13}$    |
|------|----------|--------------|----------|-------------|
| DE   | 0.79256  | 5126.497(3)  | 0.74990  | 0.08637(67) |
| PS   | 1.00040  | 5129.328(80) | 0.74990  | 0.20791(73) |
| DEPS | 1.00045  | 5126.497(1)  | 0.74990  | 0.097046    |

TABLE 4. RESULTS BY ACR#2 RULE
(WITH THE FORCING SUB-RULE)

|      | $G_3$    | $G_5$       | $G_{11}$ | $G_{13}$    |
|------|----------|-------------|----------|-------------|
| DE   | 0.79854  | 5126.500    | 0.74990  | 0.22268(6)  |
| PS   | 1.00036  | 5131.188(1) | 0.74990  | 0.097855    |
| DEPS | 1.00050  | 5126.497    | 0.74990  | 0.066257    |

Table 2 lists the results by the swarm algorithms in BCH rule, which $G_3$ and $G_5$ were also listed in [17]. Compared with table 1, it can be found that most results were not satisfied, which only DE and DEPS performed better than GA, but worse than ES for $G_5$, and swarm algorithms performed similar to ES and GA for $G_{11}$.

Table 3 lists the results by the swarm algorithms in ACR#1, which without the forcing sub-rule. Compared with the results in table 2, the results for $G_3$ and $G_{11}$ are significant better, and for $G_5$, $G_{13}$, there have more failed runs although the mean results for successful runs are better. It means that the ACR#1 has the capability for improving the performance of swarm algorithms. The results for $G_3$ and $G_{11}$, especially for PS and DEPS, seem to be comparable with that of table 1. However, the results may not enter the $S_O$, even not enter the $S_F$, due to the stagnated $\varepsilon_R^{(t)}$ in some runs.

Table 4 lists the results by the swarm algorithms in ACR#2, which with the forcing sub-rule. It can be found that the forcing sub-rule eliminated the failed runs. All the results were better than the results in BCH rule. The DE performed not so good for $G_3$ and $G_{13}$, and the PS performed not so good for $G_5$. However, the results by DEPS were better than the results by existing algorithms as listed in table 1.

TABLE 5. STANDARD DEVIATION FOR TEST PROBLEMS

|         | $G_3$     | $G_5$     | $G_{11}$ | $G_{13}$ |
|---------|-----------|-----------|----------|----------|
| ES [13] | 1.9E-04   | 3.5E+00   | 8.0E-05  | 3.1E-02  |
| GA [7]  | 7.5E-05   | 3.9E+03   | 0.0E+00  | -        |
| DEPS    | 8.12E-07  | 1.41E-10  | 0.0E+00  | 6.78E-2  |

Table 5 summaries the standard deviations of the results by ES[13], GA [7], and DEPS in ACR#2. It can also be shown that DEPS in ACR#2 has smaller standard deviations than ES[13] and GA [7], except for $G_{13}$.

It is also worthy to compare stochastic ranking (SR) technique [13] with ACR Rule, since both of them achieved good results by transforming the landscape of problem. As mentioned by Farmani and Wright [7]: "*The stochastic ranking technique has the advantage that it is simple to implement, but it can be sensitive to the value of its single control parameter.*" Besides, the SR technique is not suitable for current swarm algorithms, since it brings too much randomicity on deciding the $\vec{g}^{(t)}$ from the stochastic landscape. The ACR rule avoids such problem, since the landscape by it is fixed in one learning cycle, although it may be varied along with the learing cycles.

### VIII. Conclusions

Swarm algorithms are behavioral models based on the concepts of social swarm. Each agent in the swarm is worked by executing the action rules, which provide a self-adaptive bell-shaped variations with consensus on the diversity of swarm, in iterated learning cycles.

However, the analysis shows such self-adaption mechanism may be hard for the $S_F$ of the problems with eqaulity constraints that in ridge function class.

The adaptive constraints relaxing (ACR) rule is presented for swarm algorithms to handle with the problems with eqaulity constraints. The relaxed quasi feasible space is introduced and shrinked adaptively, so as to enter the solution space at the end of evolution. The experimental results on benchmark functions is compared with the performance of other algorithms, which shows it can achieve better results in less evalution times.